\documentclass[runningheads]{llncs}

\usepackage[T1]{fontenc}

\usepackage{graphicx}

\usepackage{hyperref}
\usepackage{color}

\usepackage{amssymb}
\usepackage{float}
\usepackage{multirow}
\usepackage{makecell}
\usepackage{lipsum}
\usepackage{bm}

\begin{document}

\title{Probabilistic Demand Forecasting with Graph Neural Networks}
\titlerunning{Probabilistic Demand Forecasting with GNNs}

\author{Nikita Kozodoi \inst{1}
    \and
    Elizaveta Zinovyeva \inst{1}
    \and
    Simon Valentin \inst{1,2}
    \and
    Jo\~{a}o Pereira \inst{3}
    \and
    Rodrigo Agundez \inst{3}
}
\authorrunning{Kozodoi et al.}

\institute{Amazon Web Services, Germany \\
    \email{\{kozodoi,zinovyee\}@amazon.com}
    \and
    University of Edinburgh, United Kingdom \\
    \email{s.valentin@ed.ac.uk}
    \and
    adidas, Amsterdam, The Netherlands \\
    \email{\{joao.pereira2, rodrigo.agundez\}@adidas.com}
    \\
}

\maketitle

\begin{abstract}
Demand forecasting is a prominent business use case that allows retailers to optimize inventory planning, logistics, and core business decisions. One of the key challenges in demand forecasting is accounting for relationships and interactions between articles. Most modern forecasting approaches provide independent article-level predictions that do not consider the impact of related articles. Recent research has attempted addressing this challenge using Graph Neural Networks (GNNs) and showed promising results. This paper builds on previous research on GNNs and makes two contributions. First, we integrate a GNN encoder into a state-of-the-art DeepAR model. The combined model produces probabilistic forecasts, which are crucial for decision-making under uncertainty. Second, we propose to build graphs using article attribute similarity, which avoids reliance on a pre-defined graph structure. Experiments on three real-world datasets show that the proposed approach consistently outperforms non-graph benchmarks. We also show that our approach produces article embeddings that encode article similarity and demand dynamics and are useful for other downstream business tasks beyond forecasting.

\end{abstract}

\let\thefootnote\relax\footnotetext{Preprint of the paper accepted to ECML PKDD 2023 ML4ITS Workshop}

%
%

%
%

\section{Introduction}
\label{sec_introduction}

Demand forecasting is a prominent business task faced by retailers. Predictions of future demand for articles sold by a retailer serve as inputs for many operational decisions, including inventory planning, logistics, and supply chain optimization \cite{Huber2020}. The importance of demand forecasting is further emphasized by a rapid growth of the e-commerce retail sector, which is expected to account for more than $\$6.51$ trillion in 2023, taking up $22.3\%$ of the global retail market \cite{Shopify2022}. Consequences of wrong demand estimates vary from missed sales opportunities due to articles being out-of-stock \cite{Fisher1996} to overproduction waste \cite{Darlington2007}.

E-commerce retailers frequently operate with thousands of articles sold simultaneously, and experience dynamic and volatile demand \cite{Gandhi2021}. This poses challenges for accurate demand forecasting. One of the key difficulties is that the article's demand depends not only on its historical demand, but also on demand for similar articles, introduction and removal of competing articles, out-of-stock events, and other information. This emphasizes the importance of incorporating article relationships into a forecasting model in a principled manner.

Traditional forecasting techniques such as Autoregressive Integrated Moving Average (ARIMA) models \cite{Box1964} are univariate and consider individual time series in isolation. In contrast, modern techniques based on deep learning models such as DeepAR \cite{Salinas2020} or Temporal Fusion Transformer \cite{Lim2021b} are trained on multiple time series simultaneously, which allows learning the past behavior across similar series. However, such global models are typically unable to incorporate inter-series relationships during inference, failing to account for the impact of related articles on the article's of interest demand. Integrating time series relationships during both training and inference is feasible with multivariate models such as Vector Autoregressive (VAR) models \cite{Zivot2006}. At the same time, such models experience difficulties with scaling to high-dimensional e-commerce environments.

Recent research has suggested using graphs to account for the inter-series correlations. Representing the data as a graph and using Graph Neural Networks (GNNs) to extract patterns from graph-structured data allows integrating time series relationships directly into a prediction model \cite{Velickovic2023}. Studies outside of the retail domain 
have developed architectures that combine recurrent models and GNNs and have shown promising results in forecasting road traffic or weather events \cite{Lira2022,Wu2019,Yu2017}. At the same time, the literature on GNNs for demand forecasting is scarce. Recently, Gandhi et al. \cite{Gandhi2021} proposed a GNN-based forecasting approach suited for a multiple-seller marketplace setting and demonstrated that GNNs have potential to improve the forecasting accuracy for cold-start articles. 

This paper proposes a novel demand forecasting approach that builds on the framework of Gandhi et al. \cite{Gandhi2021} and makes two contributions. First, we develop an end-to-end forecasting model architecture that combines two components: (i) GNN encoder that incorporates article relationships during training and inference; (ii) state-of-the-art DeepAR decoder for demand forecasting. In contrast to recurrent architectures commonly used in the literature, DeepAR produces probabilistic demand predictions, which are crucial for decision-making under uncertainty \cite{Salinas2020}. We denote the proposed forecasting model as GraphDeepAR.

Our second contribution is 
proposing a generic graph construction approach that does not require a pre-defined graph structure. We build graphs using pairwise article attribute similarity to define connections between articles, which allows leveraging article meta-data to model article relationships and augmenting it with domain knowledge. Unlike some of the previous approaches, the proposed solution is highly scalable. We test GraphDeepAR on three real-world datasets and show that it outperforms the standard DeepAR model.

The rest of the paper is organized as follows. Section \ref{sec_related_work} overviews related work on forecasting and graph machine learning. The proposed demand forecasting approach is introduced in Section \ref{sec_methods}. Section \ref{sec_setup} describes the experimental setup, whereas Section 5 provides and discusses the empirical results. Section \ref{sec_conclusion} summarizes the main conclusions taken from the experimental results.
%
%

\section{Related Work}
\label{sec_related_work}

\subsection{Leveraging Inter-Series Relationships}

Forecasting refers to prediction of future events based on the previously observed data. One of the key challenges in forecasting is accounting for relationships between multiple time series \cite{Lim2021a}. Many of the modern forecasting techniques focus on a univariate setting, where the task is limited to modeling a single time series or a small number of individual unrelated series. In practice, one is often required to forecast a large number of related series at the same time (e.g., energy consumption across different households or consumer demand for multiple related articles) \cite{Salinas2020}. In such settings, demand for a given article depends not only on its historical demand, but also on the demand for substitute and complement articles, launch of new competing articles, and other related factors \cite{Gandhi2021}.

One way to account for such relationships during training is to use global models such as DeepAR \cite{Salinas2020}. Here, we refer to models that are trained simultaneously on all time series available in a dataset as global models. Such global models are able to learn the past behavior across similar sequences. However, during inference, they are still limited to univariate forecasts, which prohibits using the previous values of related time series to predict the target series.

To incorporate inter-series relationships during both training and inference, previous work suggested multivariate forecasting models such as VAR \cite{Zivot2006} or LSTNet \cite{Lai2018}. Multivariate models use observations from all available time series as input and produce joint predictions of multiple time series \cite{benidis2022deep}. Yet, such models are typically limited to a small number of time series and do not scale well to environments with thousands of related series. Other attempts to account for series relationships rely on lower-dimensional representations such as matrix factorization techniques that scale to larger volumes of data more favorably \cite{Sen2019}.


\subsection{Forecasting with Graph Neural Networks}

Recent developments in graph machine learning have inspired researchers to explore using graphs for modeling relationships between time series \cite{Velickovic2023}. In graph machine learning, the input data is transformed into a graph, and the entities and their relationships are represented as graph nodes and edges containing a set of features. The literature has suggested GNN architectures such as Graph Convolutional Networks (GCN) to learn from graph-structured data \cite{Kipf2016}.

Prior work on GNNs for forecasting has mainly focused on traffic forecasting, where a graph representation is constructed using locations of the traffic sensors. For instance,
Yu et al. proposed the Spatio-Temporal GCN (ST-GCN) model that enables faster training \cite{Yu2017}. More recently, Wu et al. developed the GraphWaveNet architecture that learns a self-adaptive adjacency matrix and was shown to outperform DCRNN and ST-GCN \cite{Wu2019}. Other applications of GNN-based methods include forecasting frost incidence across multiple weather stations \cite{Lira2022} and passenger demand prediction across city areas \cite{Bai2019}. Going beyond applications with a clear pre-defined graph structure, Cao et al. developed a Spectral Temporal GNN (StemGNN) that uses self-attention to automatically learn latent correlations between related time series \cite{Cao2020}. However, the application of StemGNN is limited to problems with a small number of time series (i.e., less than a thousand) due to the computational cost. 

In the retail demand forecasting space, where one is faced with thousands of related time series, the literature on using GNNs remains rather limited. Gandhi et al. suggested a graph-based model that combines GNN blocks with an LSTM layer \cite{Gandhi2021}. The study has shown the superior performance of graph-based architectures compared to a non-graph baseline. At the same time, the approach proposed by Gandhi et al. is designed for a multiple-seller marketplace setting, where the graph structure is defined by seller-article relationships, and is limited to point-based predictions. This paper aims at bridging this gap and proposes a novel demand forecasting approach that extends the framework of Gandhi et al. and addresses some of the key limitations described above.

%
%

\section{Methodology}
\label{sec_methods}


\subsection{Problem Formulation}


We formalize the demand forecasting task as follows. Let $\mathcal{A} := \{ X_i, Z_i, Y_i \}_{i=1}^N$ denote the time series and features of $N$ articles sold by a retailer, where each element $A_i \in \mathcal{A}$ is a tuple with three components. $Y_i = (y_i^{t})_{t=1}^T$ is the individual time series of the $i-$th article up to the final time step $T$, such that $y_i^{t} \in \mathbb{R}^+$ represents the demand for article $i$ at some time step $t$. The articles are described with two sets of real-valued feature representations: $X_i \in \mathbb{R}^M$ is a vector with $M$ static features of article $i$, whereas $Z_i = (Z_i^t)_{t=1}^T$ with $Z_i^t \in \mathbb{R}^{L \times T}$ is a matrix with $L$ time-varying features of article $i$ at time $t$. The static features reflect article attributes such as size, color, and others. The time-varying features describe seasonal events such as week of the year, month, article promotions, and other attributes that are known in advance.

Our task is to predict demand for $N$ articles during the future $K$ time steps $(y^{T+1}_i, y_i^{T+2}, \ldots, y_i^{T+K})_{i=1}^N$ given the historical demand $(y_i^{1}, y_i^{2}, \ldots, y_i^{T})_{i=1}^N$, the static features $(X_i)_{i=1}^N$ and the time-varying features $(Z_i^{1}, Z_i^{2}, \ldots, Z_i^{T+K})_{i=1}^N$. The accuracy of the forecast is evaluated by comparing predicted demand values $(\hat{y}_i^t)_{i=1}^N$ to the actual values $(y_i^t)_{i=1}^N$ across all predicted articles and time steps.


\subsection{Graph Construction}

We address the demand forecasting task with a novel approach that splits into two components: (i) a spatial component, which includes the article graph and the GNN modules that extract knowledge from the graph-structured data; (ii) a temporal component, which is a sequence model that is able to learn from the temporal dynamics in the input time series and forecast future demand values. 

The spatial component uses a graph, where each article is represented as a node, and edges between nodes reflect article relationships. That is, we construct a graph $\mathcal{G}^t = (\mathcal{N}^t, \mathcal{E})$ that describes articles at time $t$. The graph consists of the set of nodes $\mathcal{N}^t$ that includes time-varying node features of articles from $\mathcal{A}$ that are sold at time $t$. The set of edges $\mathcal{E}$ provides pairwise node connections and contains static edge features representing article relationships.

As reported in Section \ref{sec_related_work}, graph edges are usually pre-defined by the nature of the problem. However, such knowledge is not available in a generic demand forecasting setting, which implies that relationships need to be identified by domain experts or inferred from data. This work uses attribute-based article similarity to identify article relationships automatically. We calculate pairwise cosine similarity scores between articles using static article features: 
\begin{equation}
\textrm{\mbox{similarity}}\left(A_i, A_j\right):=\cos\left(X_i, X_j\right)= \frac{X_i\cdot X_j}{||X_i|| ||X_j||}
\end{equation}
Note that calculating pairwise similarity values has exponential time complexity. At the same time, this calculation is only required once on the graph construction stage, and the graph is stored and reused for different model configurations. To optimize the time and memory usage of the calculation, we split the data in article chunks that can be run in parallel to construct the full similarity matrix.

We define an edge between two articles in $\mathcal{E}$ if their similarity exceeds a specified threshold, which serves as a graph construction hyperparameter. This approach allows us to incorporate domain expertise using the article features but also enables automatic identification of edges controlled by the choice of graph hyperparameters. 
All edges in the constructed graph are undirected.


The node features in $\mathcal{N}^t$ contain article information leveraged by the model to make demand forecasts. In our setting, an article can be described by the attribute-based features and the previous demand lags (i.e., demand across the previous timestamps). Since article attributes are static and can be directly inputted in the temporal component, we limit node features to $P$ demand lags: $\mathcal{N}_i^t = (y_i^{t-P+1}, y_i^{t-P+2}, \ldots, y_i^{t})$. The number of lags serves as a hyperparameter. In addition, we calculate the node in-degree, which measures the number of neighbors of an article and serves as an additional node feature included in $\mathcal{N}_i^t$.


\subsection{Model Architecture}

\begin{figure}\centering
\includegraphics[trim={0 0.77cm 0 0.10cm},width=\textwidth,clip]{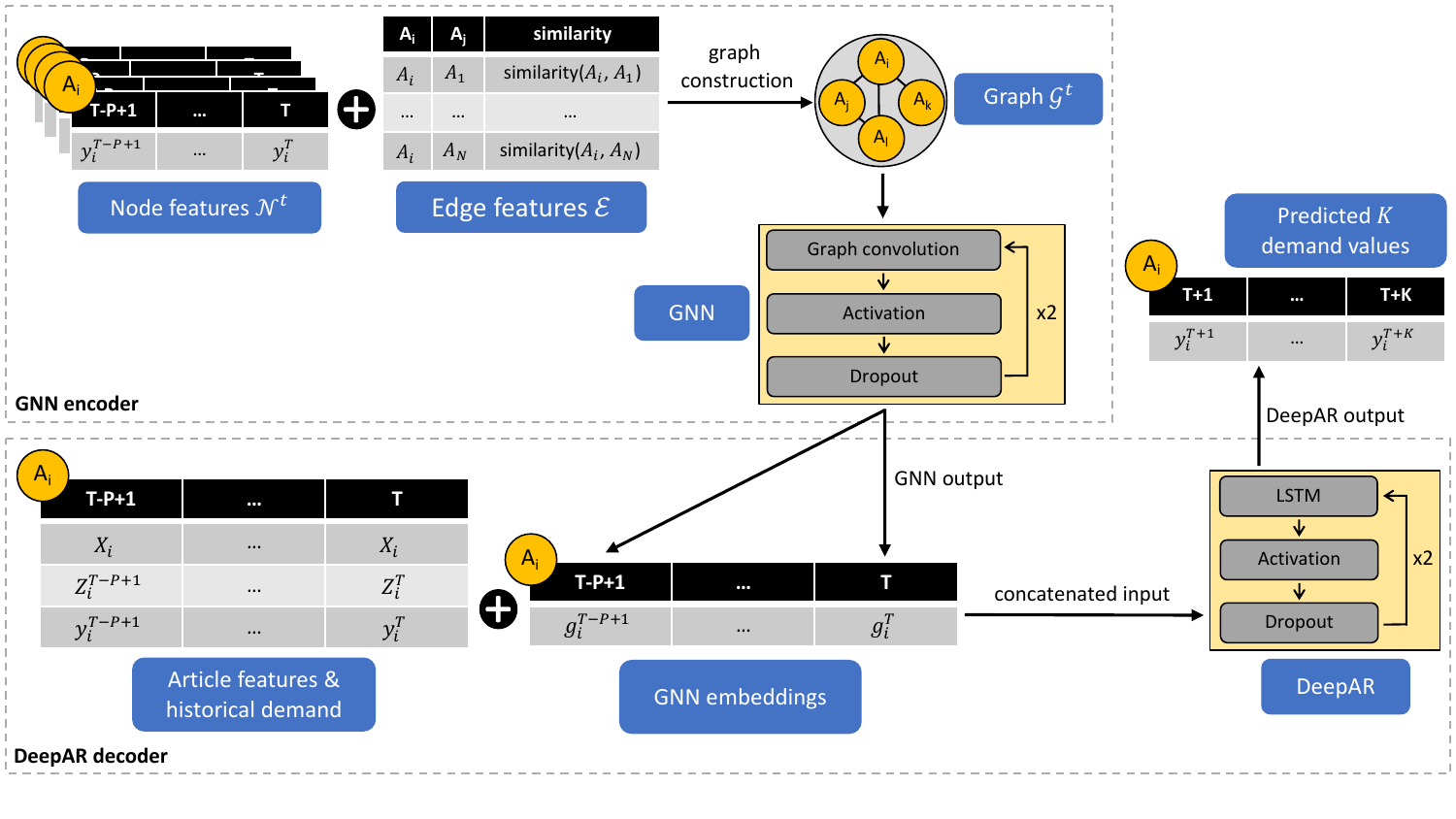}
\caption{GraphDeepAR Architecture. The figure uses a single article for illustration.} \label{fig_model}
\end{figure}

The proposed demand forecasting approach leverages a deep neural network that combines a GNN encoder and a DeepAR decoder. Both components are trained end-to-end with a single loss function. Further, we refer to the end-to-end model as GraphDeepAR. The model architecture is depicted in Figure \ref{fig_model}.

GraphDeepAR processes articles in mini-batches. For every batch presented to the model, we construct a graph with dynamic node features containing the article demand values over a certain time window, which spans over the $P$ consecutive time steps prior to the last timestamp in the mini-batch. The similarity-based graph edges remain static over the timestamps, whereas the node features are calculated based on the timestamp of the corresponding graph.

The graph $\mathcal{G}^t = (\mathcal{N}^t, \mathcal{E})$ serves as input for the two-layer GNN encoder. Each of the two layer blocks consists of a graph-based neighborhood pooling, activation function and dropout. The graph convolution layers represent article $i$ using aggregated node features of the neighboring articles $\mathcal{N}_{j \in I}^t$, where $\mathcal{I}$ is a set of articles connected with article $i$. This allows the encoder to gather information from similar articles and output article embeddings that encode information in its neighborhood. Formally, the output of the $k$-th GNN layer for article $i$ at time $t$ is denoted as $h^t_k(i)$ and calculated as follows: 
\begin{equation}
h_{k(i)}^t = \sigma \left( \displaystyle \frac{1}{|\mathcal{N}_j^t|} \ \sum\limits_{j \in \mathcal{I}} W_k h_{(k-1)}^t(j) \right),
\end{equation}
where $W$ is a weight matrix to be learned from data, and $\sigma(\cdot)$ is a Leaky ReLU activation function \cite{maas2013rectifier}. GNN layers share parameters in $W$ across the timestamps, which implies that the encoder iteratively learns how to leverage node features irrespective of the time step. The embeddings generated by the last GNN layer are denoted as $G_i = (g_i^{t-P+1}, g_i^{t-P+2} \ldots, g_i^t)$ with $G_i \in \mathbb{R}^{D \times P}$, where $D$ is a hyperparameter indicating the embedding dimensionality.

The temporal component represents a two-layer DeepAR forecasting model that produces demand predictions for each article. During training and inference, GNN embeddings $(G_i)_{i=1}^N$ encoding the data from the article neighborhood are concatenated with previous demand values $(y_i)_{i=1}^N$ that both cover the time window of length $P$, static article features $(X_i)_{i=1}^N$, and known time-varying features $(Z_i)_{i=1}^N$ for the previous $P$ and the following $K$ time steps. The concatenated matrix serves as input to the DeepAR decoder. The decoder outputs the mean $\mu$ and variance $s^2$ of a Students' $t$-distribution that is used to model a distribution over plausible future demand values. This choice is motivated by the heavier tails of the Student's $t$-distribution as compared to a Gaussian, which is more appropriate for typical demand dynamics in the retail domain. The demand predictions for the following $K$ time steps are then sampled from $t(\mu, s)$.
%
%

\section{Experimental Setup}
\label{sec_setup}


\subsection{Data}

Table \ref{tab_data} overviews the datasets used in the paper. Each dataset provides demand across different articles sold by a retailer. The first two datasets, \emph{retail} and \emph{e-commerce}, are publicly available on Kaggle\footnote{\scriptsize{Source: \url{https://www.kaggle.com/datasets/berkayalan/retail-sales-data}, \url{https://www.kaggle.com/competitions/competitive-data-science-predict-future-sales/data}}}. The \emph{adidas} dataset \emph{(hidden for review)} is proprietary sales data provided by adidas. The number of articles is ranging between $629$ and 80,838, which allows us to evaluate our approach across environments with different dimensionality.

In each dataset, we aggregate weekly demand across all stores to produce the target time series and use between $5$ and $20$ features, including static features describing the article attributes (e.g., size and category), and three dynamic features describing the seasonal patterns (week of the year, day of the month, week number). Since the original \textit{e-commerce} dataset includes many articles where demand is low and static, we limit the training set for this dataset to articles that have the standard deviation of the weekly demand greater than $1$.

We split each dataset into three subsets along the time axis: training, validation, and test sets. The test set is used for model evaluation and covers the last $26$ weeks. The validation set is used to tune model hyperparameters; the length of the validation set is fixed to $13$ weeks preceding the test set. The training set covers all data available prior to the validation set. Depending on data availability, each article may be observed in all three data subsets.

\begin{table}\centering    
    \caption{Datasets Summary}
    \label{tab_data}
    \begin{tabular}{@{\extracolsep{10pt}} lccc}
    \hline
    Dataset & No. articles & No. weeks & No. features \\
    \hline
    Retail     & 629    & 148  & 12 \\
    E-commerce & 8,810  & 128  & 5  \\
    adidas     & 80,838 & 140  & 20 \\ 
    \hline
\end{tabular}
\end{table}


\subsection{Setup}

To evaluate our approach, we compare the proposed GraphDeepAR model to a DeepAR benchmark. The two models are trained and evaluated on the same data splits and share the same hyperparameters. For example, for the datasets \emph{retail} and \emph{e-commerce}, both models use two LSTM layers with a hidden size of $128$ and are trained to minimize the $t$-distribution loss with a symmetric loss penalty. For the \emph{adidas} dataset, we use an asymmetric penalty to treat differently under- and over-prediction costs based on the business context. Appendix A reports hyperparameter values used on the two public datasets.

There are two differences between DeepAR and GraphDeepAR related to the addition of the graph component. First, GraphDeepAR includes a GNN encoder described in Section \ref{sec_methods}. For efficient training, the GNN encoder performs neighborhood sampling that limits the maximum number of article neighbors considered on each training pass. This is achieved by removing a random subset of edges for each node in the graph, similar to the sampling strategy introduced in GraphSAGE \cite{hamilton2017inductive}. Second, while DeepAR benefits from randomized batching, GraphDeepAR uses synchronized batching, which implies that all demand sequences inside one mini-batch cover the same time window. Synchronized batching reduces the randomness in data shuffling but is required for GraphDeepAR to ensure that batch-level graphs have node features that are consistent in time.


The forecasting accuracy is evaluated using multiple metrics. On the public datasets, 
we use three established forecasting metrics: Root Mean Squared Error (RMSE), Mean Absolute Error (MAE), and Weighted Mean Absolute Percentage Error (WMAPE), which uses actual demand as weights to assign a higher importance to top-selling articles. The \emph{adidas} dataset provides the opportunity to directly measure the forecast quality from a business perspective based on the financial value of under- and over-prediction. Here, we calculate the monetary loss of two models and report the financial uplift from GraphDeepAR as a percentage improvement relative to the DeepAR baseline.

Both models are implemented using the PyTorch Forecasting library \cite{Pytorch_Forecasting}. We also use the Deep Graph Library \cite{wang2019dgl} to construct the graph and implement the GNN encoder. Training is performed in PyTorch Lightning \cite{Falcon_PyTorch_Lightning}. All experiments are conducted on compute instances running on Amazon SageMaker.
%
%

\section{Results}
\label{sec_results}


\subsection{Graph Illustration}

\begin{figure}\centering
\includegraphics[trim={4cm 4.6cm 4cm 4.5cm},width = 0.44\textwidth,clip]{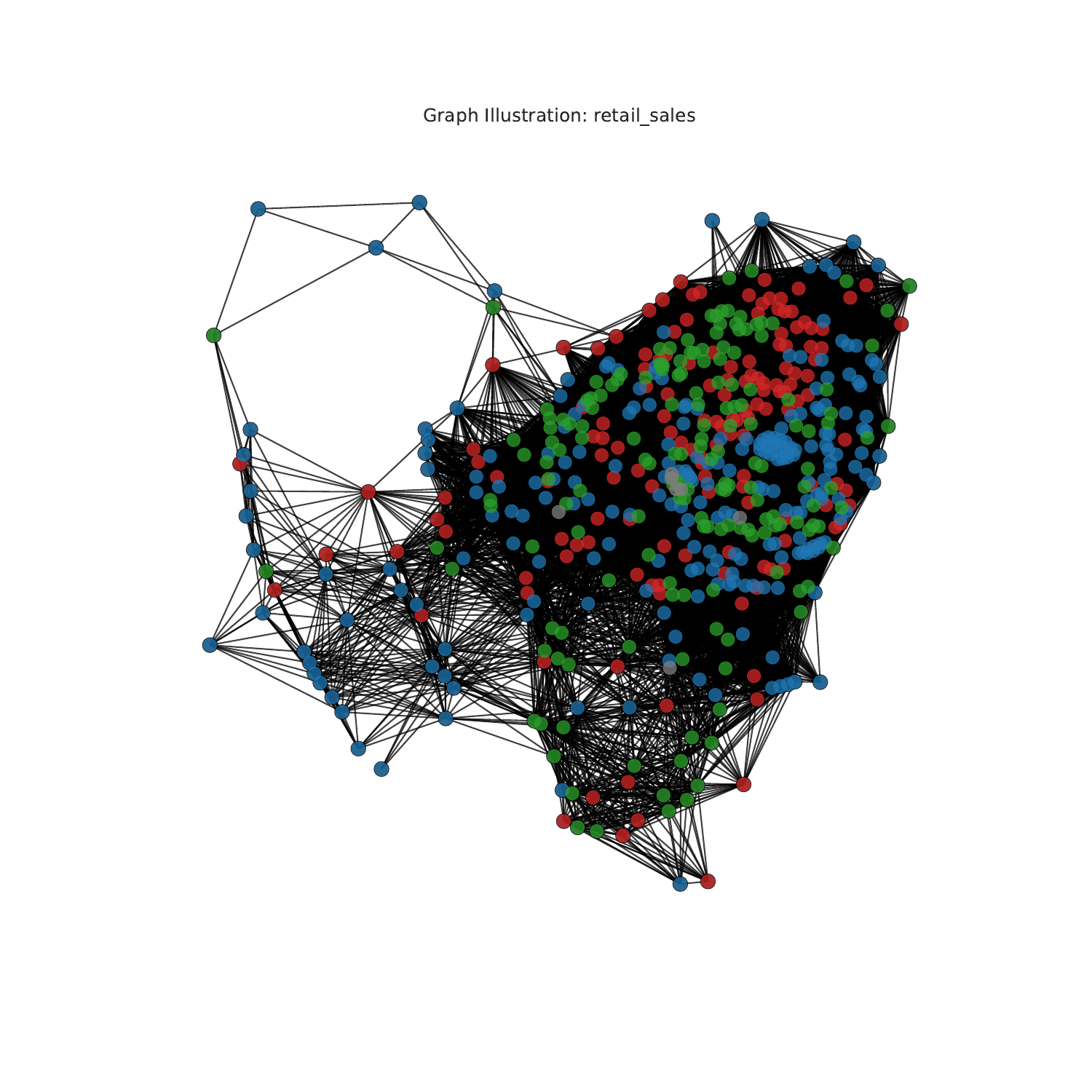}
\includegraphics[trim={4cm 4.6cm 4cm 4.5cm},width = 0.44\textwidth,clip]{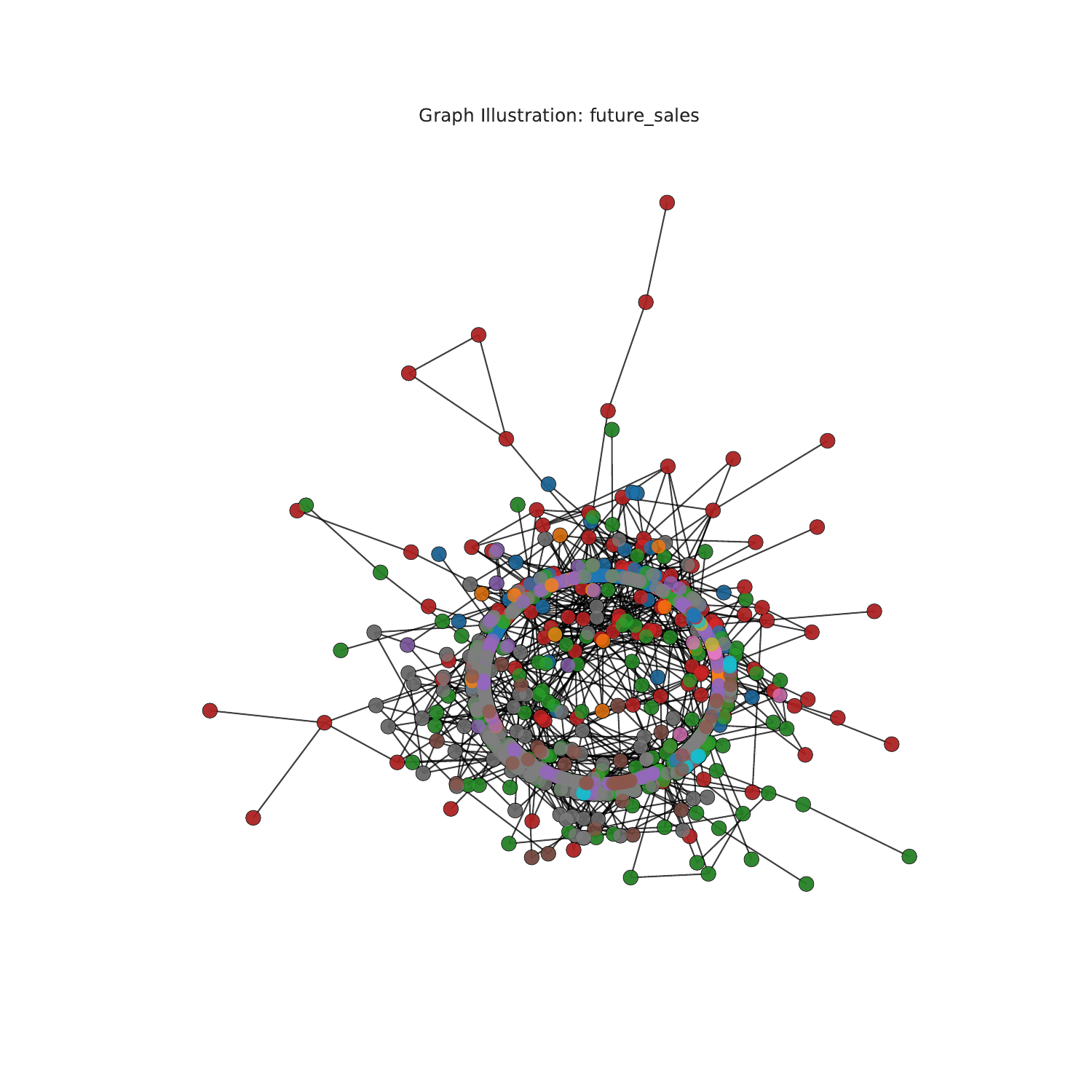}
\caption{Graph Illustration. Article graphs for \textit{retail} (left) and \textit{e-commerce} (right).} \label{fig_graph}
\end{figure}

Figure \ref{fig_graph} illustrates graphs constructed for the two public datasets. Each node represents an article color-coded with the article category. For illustration purposes, we do not display the self-connecting edges and limit the \emph{e-commerce} graph depicted in the right panel to 1,000 top-selling articles. 

As described in Section \ref{sec_methods}, graph edges are based on article similarity. The \emph{retail} dataset provides nine article features used to calculate pairwise cosine distance. The \emph{e-commerce} data only includes two categorical features indicating the article hierarchy. To address this data scarcity, we encode article titles using a sentence transformer-based encoder with distilBERT \cite{reimers-2019-sentence-bert,sanh2019distilbert}. 
The produced article embeddings of dimension $512$ are used to calculate the cosine similarity.

As shown in Figure \ref{fig_graph}, the neighborhood size varies considerably across articles: both graphs contain clusters of densely connected articles as well as articles with no neighbors ($42\%$ in the \emph{e-commerce} graph and $3\%$ -- in \emph{retail}). Given the $0.95$ similarity threshold, the average number of neighbors is $1.21$ in the \emph{e-commerce} graph and $198.13$ in the \emph{retail} graph (with a standard deviation of $114$). This emphasizes the importance of selecting a suitable similarity threshold and using neighborhood sampling to facilitate fast training. Such variations in graph structure may affect forecasting accuracy differently for different articles, depending on the article's position in the graph and the number of neighbors.


\subsection{Predictive Performance}
\label{sec_results_perf}

Table \ref{tab_res_public} presents empirical results on two public datasets. Apart from the overall forecasting accuracy across all articles, we calculate metrics across three groups: \textit{cold starts, connected articles,} and \textit{Top-$100$ articles} -- the top-selling articles with the highest total demand. We define cold starts as articles with less than five historical demand values at the time of the forecast, and connected articles as articles that have edge connections to other articles in the graph. Examples of individual demand predictions for several articles are illustrated in Appendix B.

\begin{table}\centering    
    \caption{Results: Retail and E-Commerce Datasets}
    \label{tab_res_public}
    \begin{tabular}{@{\extracolsep{8pt}} lllccc}
    \hline
    Dataset & Subset & Model & RMSE & MAE & WMAPE \\
    \hline
    \multirow{8}{*}{Retail} & \multirow{2}{*}{All articles}     & DeepAR      & 204.68 & 51.53 & 0.43 \\
                            &                                   & GraphDeepAR & \textbf{196.13} & \textbf{50.35}  & \textbf{0.42} \\
    \cline{2-6}
                            & \multirow{2}{*}{Cold starts}  & DeepAR      & 44.79 & 19.83  & 0.66 \\
                            &                               & GraphDeepAR & \textbf{41.78} & \textbf{18.84}  & \textbf{0.63} \\
    \cline{2-6}
                            & \multirow{2}{*}{Connected articles} & DeepAR      & 207.12 & 52.34  & 0.42 \\
                            &                                            & GraphDeepAR & \textbf{198.46} & \textbf{51.12} & \textbf{0.41} \\
    \cline{2-6}
                            & \multirow{2}{*}{Top-100 articles} & DeepAR      & 419.40 & 171.28  & 0.36 \\
                            &                                   & GraphDeepAR & \textbf{401.27} & \textbf{164.10}  & \textbf{0.35} \\
    \hline
    \multirow{8}{*}{E-commerce} & \multirow{2}{*}{All articles} & DeepAR      & 30.36 & 3.39 & 0.67 \\
                                &                               & GraphDeepAR & \textbf{20.65} & \textbf{3.08} & \textbf{0.61} \\
    \cline{2-6}
                                & \multirow{2}{*}{Cold starts}      & DeepAR      & \textbf{8.66} & 2.62 & 0.79 \\
                                &                                   & GraphDeepAR & 8.72 & \textbf{2.62} & \textbf{0.79} \\
    \cline{2-6}
                            & \multirow{2}{*}{Connected articles} & DeepAR      & 31.40 & 3.59 & 0.69 \\
                            &                                     & GraphDeepAR & \textbf{21.40} & \textbf{3.18} & \textbf{0.61} \\
    \cline{2-6}    
                                & \multirow{2}{*}{Top-100 articles} & DeepAR      & 164.68 & 42.78 & 0.98 \\
                                &                                   & GraphDeepAR & \textbf{110.50} & \textbf{29.60} & \textbf{0.68} \\
    \hline
\end{tabular}
\end{table}

Comparing the overall results, we see that GraphDeepAR consistently outperforms DeepAR in all three evaluation metrics. The largest gains are observed for the RMSE metric, where GraphDeepAR outperforms DeepAR by $4.36\%$ on \textit{retail} and by $31.98\%$ on \textit{e-commerce} data. The bigger improvement on \textit{e-commerce} can be explained by a substantially larger sample size of the dataset.

Performance gains vary across the article groups. On the \textit{retail} dataset, the largest RMSE uplift of $6.72\%$ is observed for cold starts, which suggests that leveraging historical demand for similar articles is particularly useful if there are no or too few previous demand values for a given article. Interestingly, we do not observe improvements for cold start articles on \textit{e-commerce} data. This can be explained by the fact that cold starts in this dataset are rare ($2\%$ of the considered articles), have $30\%$ less neighbors compared to carry-over articles, and tend to have a lower and more stable weekly demand with a standard deviation of $2.29$. This is also reflected in smaller absolute values of MAE and RMSE achieved by both forecasting models for cold start articles on the \textit{e-commerce} dataset in comparison to other article groups.

Considering further article groups, GraphDeepAR consistently outperforms DeepAR on connected articles on both datasets. This result is in line with our expectations, since GraphDeepAR aggregates historical demand values over the article neighborhood, which helps to improve the forecast performance for well-connected articles. Articles with neighbors constitute $97\%$ articles on \textit{retail} data and $58\%$ articles on \textit{e-commerce} data, which corresponds to a large share of the sold articles and emphasizes the importance of the observed gains.

The lack of financial data for the two public datasets makes it challenging to translate the observed gains into monetary values. As a step in this direction, we evaluate models on $100$ top-selling articles, which are likely to be stronger drivers of the retailer's profit compared to the rarely-sold articles. The results indicate that GraphDeepAR consistently improves the forecasting performance for the top-selling articles, achieving a $4.32\%$ RMSE uplift on \textit{retail} data and a $32.90\%$ uplift on the \textit{e-commerce} dataset.

\begin{table}\centering    
    \caption{Results: adids Dataset}
    \label{tab_res_adidas}
    \begin{tabular}{@{\extracolsep{10pt}} llc}
    \hline
    Dataset & Subset & Financial uplift \\
    \hline
    \multirow{7}{*}{adidas} & Test set 1 & $5.42\%$ \\
                            & Test set 2 & $4.05\%$ \\
                            & Test set 3 & $1.66\%$ \\
                            & Test set 4 & $0.52\%$ \\
                            & Test set 5 & $0.33\%$ \\
                            & Test set 6 & $0.32\%$ \\
    \cline{2-3}
                            & Average & $2.05\%$ \\
    \hline
\end{tabular}
\end{table}

The \textit{adidas} dataset includes information needed to quantify the retailer's financial efficiency. Table \ref{tab_res_adidas} reports the financial uplift from GraphDeepAR across six article subsets. GraphDeepAR consistently outperforms DeepAR on all article splits, achieving the average financial uplift of $2.05\%$. This implies that performance gains from our approach are reflected not only in statistical metrics, but also in a tangible financial value improvement.


\subsection{Runtime}

Introducing a GNN encoder increases the computational complexity of the model. Table \ref{tab_time} compares the running time of DeepAR and GraphDeepAR across the three datasets. We provide the training and inference times and calculate the total running time difference as a percentage increase relative to DeepAR. Model training and inference is performed using the same Amazon SageMaker instance.

\begin{table}\centering    
    \caption{Running Time Differences}
    \label{tab_time}
    \begin{tabular}{@{\extracolsep{5pt}} llccc}
    \hline
    Dataset    & Model & Training time & Inference time & Total difference \\
    \hline
    \multirow{2}{*}{Retail}     & DeepAR      & 10.80 min  & 0.14 min  & \multirow{2}{*}{160.96\%} \\
                                & GraphDeepAR & 28.33 min  & 0.22 min  &                           \\
    \hline
    \multirow{2}{*}{E-commerce} & DeepAR      & 90.28 min  & 3.26 min & \multirow{2}{*}{154.64\%} \\
                                & GraphDeepAR & 234.73 min & 3.46 min &                         \\
    \hline
    \multirow{2}{*}{adidas} & DeepAR      & 55.92 min  & 20.69 min  & \multirow{2}{*}{120.28\%} \\
                                  & GraphDeepAR & 139.80 min  & 28.96 min  &                           \\
    \hline
\end{tabular}
\end{table}

As expected, GraphDeepAR exhibits a longer running time. The largest differences are observed during training, where GraphDeepAR is around $159\%$ slower due to the need to backpropagate gradients through additional GNN layers. 
Interestingly, inference with GraphDeepAR is only around $34\%$ slower, indicating that leveraging graph-structured data on the prediction stage requires less additional resources. Note that the absolute training times of GraphDeepAR are not prohibiting from regularly retraining the forecasting model on a weekly or monthly basis to dynamically update the demand projections. This implies that in practice, using GraphDeepAR will incur substantially higher costs only if the forecasting model has to be retrained more frequently. 

Compared to DeepAR, GraphDeepAR requires an additional data processing step that involves pairwise article similarity calculation and graph construction. In our experiments, the time required to build and export the article graph varied between $5$ and $52$ min across the three datasets. Note that this calculation only needs to be done once and can be reused for different model configurations as long as no new articles are introduced by a retailer.


\subsection{Article Embeddings}

GraphDeepAR can be used to produce time-varying article embeddings. The embeddings $(G_i)_{i=1}^N$ outputted by the GNN encoder are learned during training in an end-to-end fashion and leverage data contained in node and edge features to produce article representations useful for demand forecasts. The resulting embeddings can be used in further downstream tasks, such as pricing problem.

\begin{figure}\centering
\includegraphics[width = 0.49\textwidth,trim={0 0.9cm 0 1.5cm},clip]{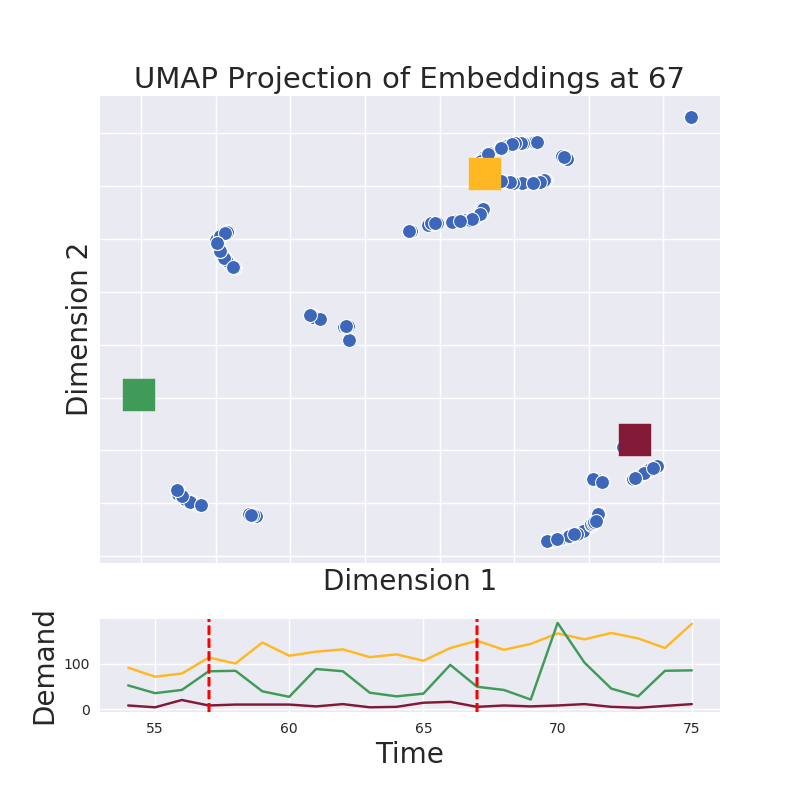}
\includegraphics[width = 0.49\textwidth,trim={0 0.9cm 0 1.5cm},clip]{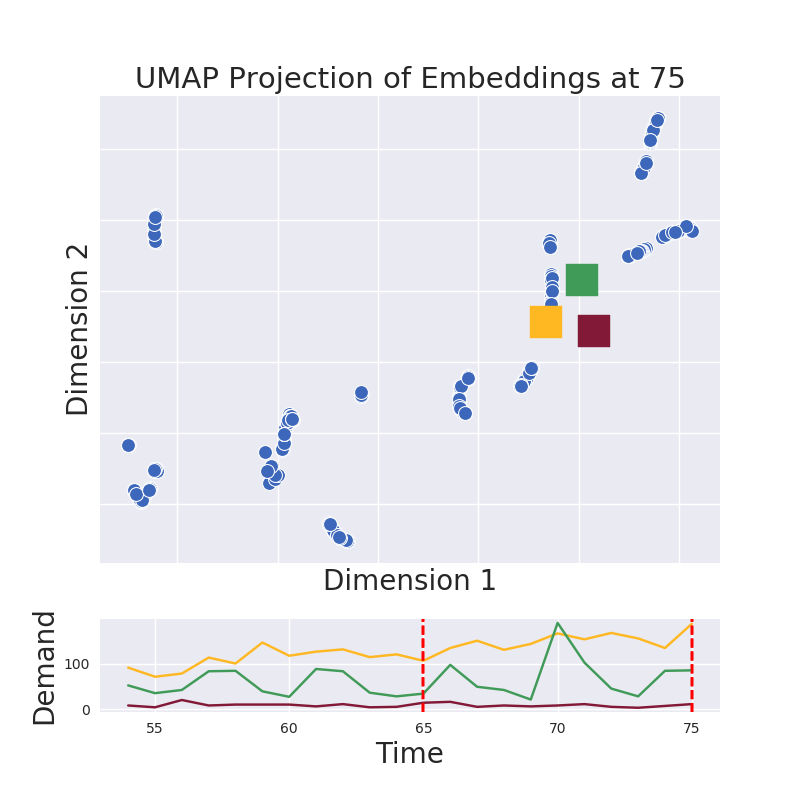}
\caption{GNN article embeddings for \textit{retail} for week 67 (left) and 75 (right).} \label{fig_embeddings}
\end{figure}

Figure \ref{fig_embeddings} illustrates GraphDeepAR's embeddings for a subset of articles from the \emph{retail} dataset for two time steps. Embeddings are mapped onto a two-dimensional space using UMAP \cite{mcinnes2018umap} and are depicted as a scatter plot in the upper part of the figure. We highlight three neighboring articles with square markers. The lower part of the figure shows demand for the highlighted articles and a box indicating the time window covered by the embedding. 

As shown in Figure \ref{fig_embeddings}, in week 67 (left panel), demand correlation between the neighboring articles is negative. Here, despite showing a high attribute-based similarity, the articles are located far away from each other in the 2D projection of the embedding space. At the same time, in week 75 (right panel), where articles' demand goes in the same direction, the articles move closer to each other. This illustrates that GNN embeddings integrate two types of information: initial article features used for graph construction and demand dynamics of the neighboring articles. This makes the embeddings useful for other prediction tasks, by serving as a proxy to indicate substitute and complement articles, under appropriate conditions.
%
%

\section{Conclusion}
\label{sec_conclusion}

This paper proposes a novel demand forecasting approach that leverages Graph Neural Networks (GNNs) to account for article relationships during training and inference. We introduce an end-to-end GraphDeepAR model that provides probabilistic demand predictions and avoids reliance on a pre-defined graph structure for graph construction. The approach is tested on three real-world datasets. The results indicate that GraphDeepAR outperforms a non-graph benchmark in statistical metrics and financial value, demonstrating the added value of integrating the article graph. We also show that GraphDeepAR produces dynamic GNN embeddings that can represent articles in different downstream prediction tasks.

Empirical comparisons provide guidelines for decision-makers, revealing domains in which retailers are more likely to benefit from a graph-based approach. Across the two public datasets, we observe higher gains on a larger dataset, which indicates that GraphDeepAR has higher potential given a large number of related articles. We also observe consistent accuracy gains on articles that have neighbors in the graph. Constructing a well-connected graph is, therefore, an important requirement, since GraphDeepAR requires edge connections to leverage patterns observed for related articles. 
The good performance of GraphDeepAR on top-selling articles indicates that our approach is able to improve forecasting accuracy on the popular articles that drive the retailer's revenue. 

The accuracy gains come at a cost of longer running times. On average, GraphDeepAR is $159\%$ slower during the training stage and $34\%$ slower during inference. This indicates that our approach incurs higher costs if the forecasting model has to be retrained frequently, whereas producing predictions with GraphDeepAR requires a reduced number of additional resources.


The proposed approach introduces additional hyperparameters into DeepAR, including graph construction (e.g., similarity measure and cutoff) and GNN encoder  (e.g., embedding sizes and neighborhood sampling ratios). The future research could perform ablation studies to investigate the impact of important hyperparameters on the model performance and training time.

Using the GNN encoder implies that demand predictions are affected by all neighboring articles, which makes GraphDeepAR useful for simulating certain scenarios. One promising direction would be to investigate how forecasts are affected by introducing or removing articles, which requires reconstructing the graph at the inference time.
Another idea concerns exploring alternative ways of modeling article relationships. For example, historical demand correlation could serve as a proxy to build connections if article attribute data is not available. 

%
%


\bibliographystyle{splncs04}
\bibliography{references}

\newpage
\section*{Supplementary Material}

\subsection*{Appendix A: Meta-Parameters}

Table \ref{tab_params} provides the used meta-parameter values of DeepAR and GraphDeepAR on \emph{retail} and \emph{e-commerce} datasets. We split meta-parameters into three groups: sequential model, GNN encoder, and the training procedure. Note that the GNN encoder meta-parameters are only relevant for the GraphDeepAR model.

\begin{table}[H]\centering    
    \caption{Model Meta-Parameters}
    \label{tab_params}
    \begin{tabular}{@{\extracolsep{5pt}} lllcc}
    \hline
    Dataset & Component & Meta-parameter & DeepAR & GraphDeepAR \\
    \hline
    \multirow{18}{*}{retail} & \multirow{5}{*}{Sequential model} & No. layers     & 2          & 2 \\
                             &                                   & Hidden size    & [128, 128] & [128, 128] \\
                             &                                   & Cell type      & LSTM       & LSTM \\
                             &                                   & Dropout        & 0.2        & 0.2 \\
                             &                                   & Context length & 10         & 10  \\
    \cline{2-5}
                             & \multirow{7}{*}{GNN encoder}      & No. layers        & -- & 2       \\
                             &                                   & Hidden size       & -- & [16, 8] \\
                             &                                   & Cell type         & -- & GCN     \\
                             &                                   & Dropout           & -- & 0.2     \\
                             &                                   & Similarity cutoff & -- & 0.95    \\
                             &                                   & Max no. neighbors & -- & 10      \\
                             &                                   & Context length    & -- & 10      \\
    \cline{2-5}
                             & \multirow{6}{*}{Training procedure}  & Max no. epochs & 50               & 50 \\
                             &                                      & Early stopping & 5                & 5 \\
                             &                                      & Learning rate  & $5\times10^{-3}$ & $5\times10^{-3}$ \\
                             &                                      & Optimizer      & Ranger           & Ranger \\
                             &                                      & Loss function  & t-distribution   & t-distribution \\
                             &                                      & Batch sampler  & Random           & Synchronized \\
    \hline
    \multirow{18}{*}{e-commerce} & \multirow{5}{*}{Sequential model} & No. layers     & 2          & 2 \\
                                 &                                   & Hidden size    & [128, 128] & [128, 128] \\
                                 &                                   & Cell type      & LSTM       & LSTM \\
                                 &                                   & Dropout        & 0.2        & 0.2 \\
                                 &                                   & Context length & 10         & 10  \\
    \cline{2-5}
                             & \multirow{7}{*}{GNN encoder}      & No. layers        & -- & 2       \\
                             &                                   & Hidden size       & -- & [16, 8] \\
                             &                                   & Cell type         & -- & GCN     \\
                             &                                   & Dropout           & -- & 0.2     \\
                             &                                   & Similarity cutoff & -- & 0.95    \\
                             &                                   & Max no. neighbors & -- & 5       \\
                             &                                   & Context length    & -- & 10      \\
    \cline{2-5}
                             & \multirow{6}{*}{Training procedure}  & Max no. epochs & 50               & 50 \\
                             &                                      & Early stopping & 5                & 5 \\
                             &                                      & Learning rate  & $1\times10^{-2}$ & $1\times10^{-2}$ \\
                             &                                      & Optimizer      & Ranger           & Ranger \\
                             &                                      & Loss function  & t-distribution   & t-distribution \\
                             &                                      & Batch sampler  & Random           & Synchronized \\
    \hline
\end{tabular}
\end{table}

\subsection*{Appendix B: GraphDeepAR Predictions}

\begin{figure}\centering
\includegraphics[trim={3cm 1.15cm 3cm 0.9cm},width = 0.95\textwidth,clip]{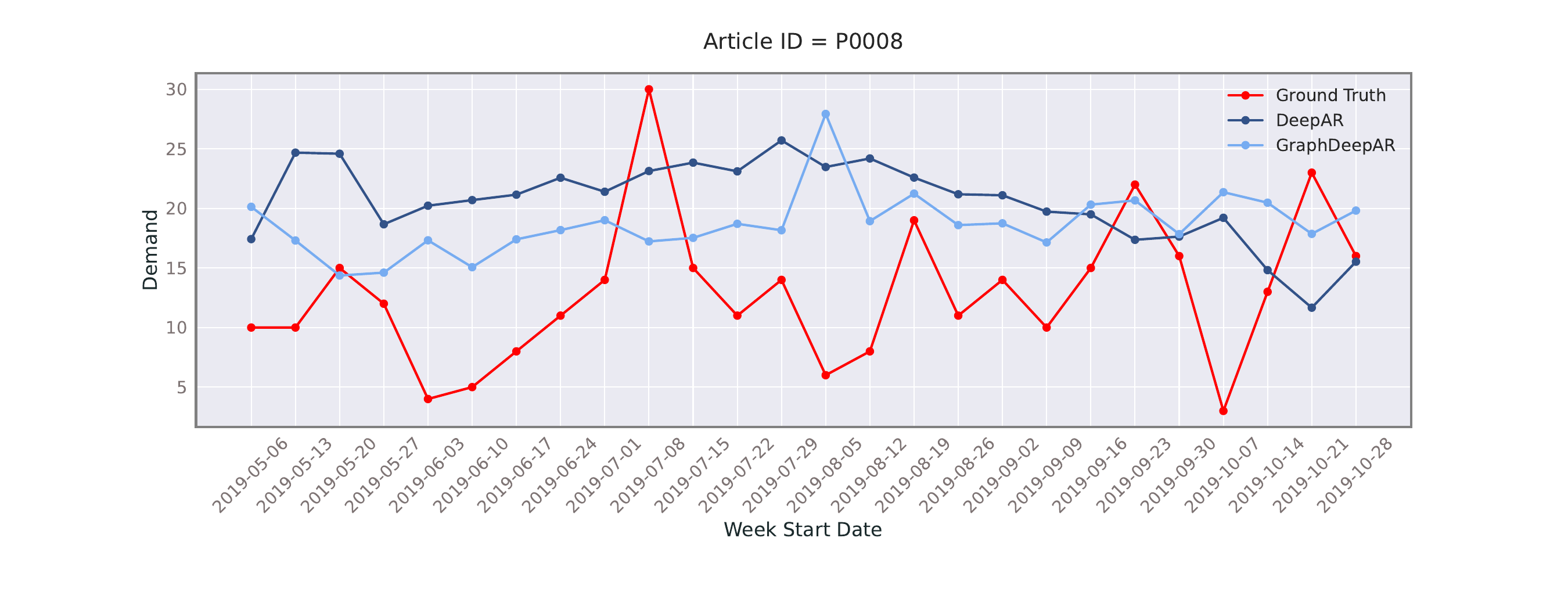}
\includegraphics[trim={3cm 1.15cm 3cm 0.9cm},width = 0.95\textwidth,clip]{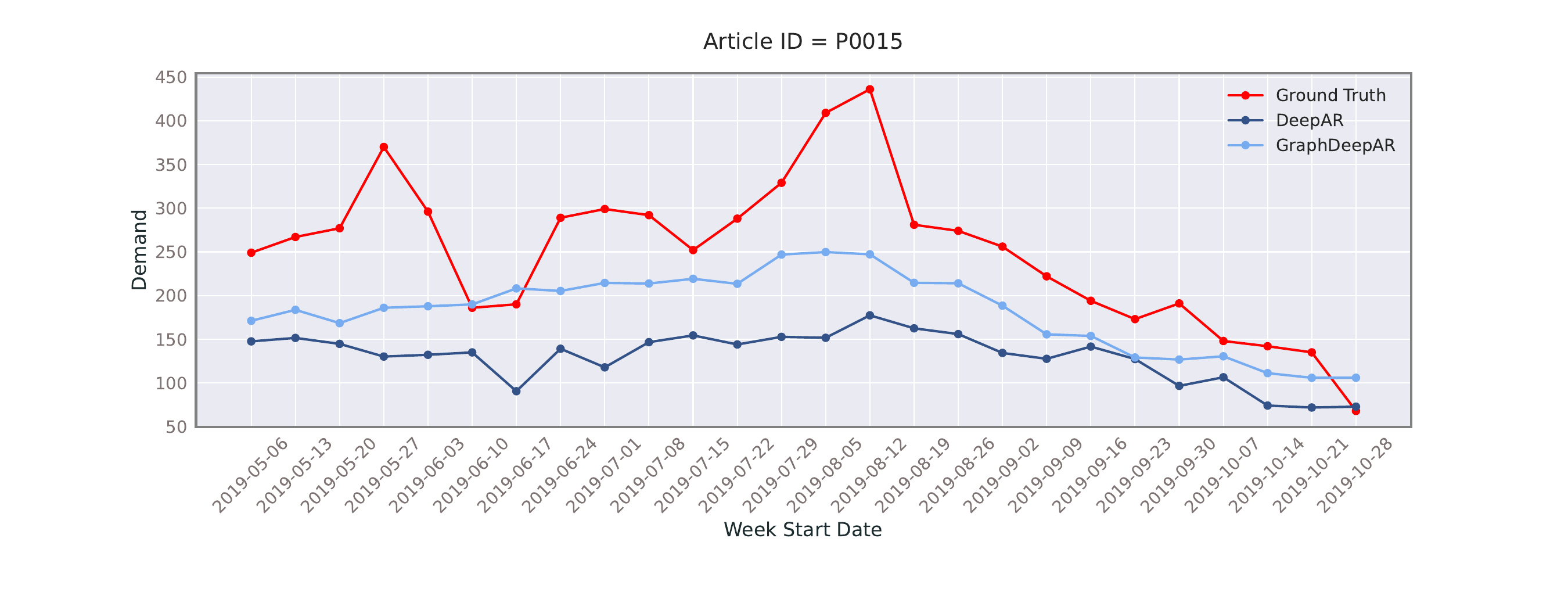}
\includegraphics[trim={3cm 1.15cm 3cm 0.9cm},width = 0.95\textwidth,clip]{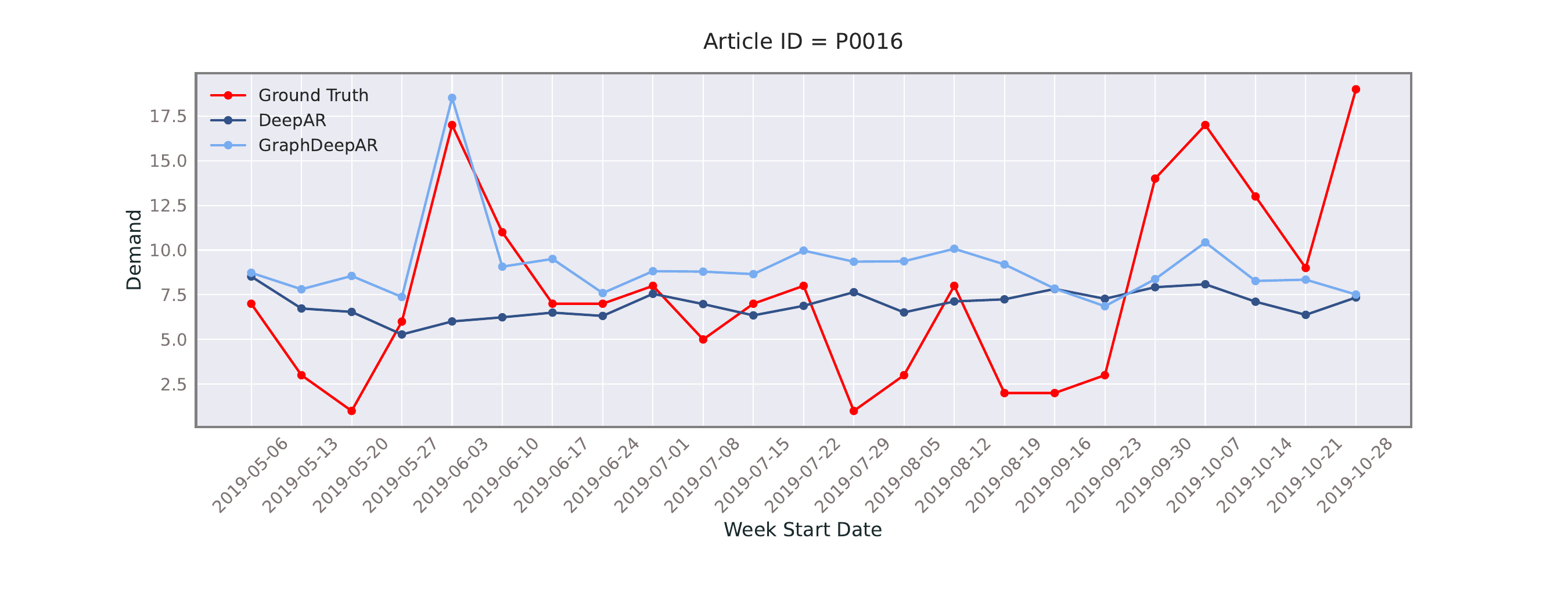}
\caption{Example predictions of DeepAR and GraphDeepAR on the \textit{Retail} dataset.} \label{fig_preds}
\end{figure}

Figure \ref{fig_preds} depicts several individual demand predictions produced by the DeepAR and GraphDeepAR models on the \emph{retail} dataset. As illustrated on the figure, GraphDeepAR is able to provide better forecasts for some of the articles.

\end{document}